\documentclass{article}
\usepackage[utf8]{inputenc} 
\usepackage[T1]{fontenc}    
\usepackage{lmodern}
\usepackage{hyperref}       
\usepackage{url}            
\usepackage{booktabs}       
\usepackage{amsfonts}       
\usepackage{nicefrac}       
\usepackage{microtype}      
\usepackage{fullpage}
\usepackage{amsmath}
\usepackage{mathrsfs}
\usepackage{graphicx}
\usepackage{dsfont}
\usepackage{wrapfig}
\usepackage{comment}
\usepackage{rotating}
\usepackage{hyperref}
\usepackage{subcaption}
\usepackage[font={small}]{caption}  
\usepackage{algorithmic}
\usepackage[ruled,linesnumbered,shortend,vlined,noend]{algorithm2e}
\usepackage[all]{xy}
\usepackage{xcolor}
\usepackage[capitalize, noabbrev]{cleveref}
\usepackage{arydshln}
\usepackage{enumitem}
\setlist[itemize]{leftmargin=*}
\usepackage{wrapfig}  

\usepackage{tabu, multirow}
\usepackage{booktabs}

\usepackage{amsmath, amssymb, amsthm}

\makeatletter
\def\thickhline{%
  \noalign{\ifnum0=`}\fi\hrule \@height \thickarrayrulewidth \futurelet
   \reserved@a\@xthickhline}
\def\@xthickhline{\ifx\reserved@a\thickhline
               \vskip\doublerulesep
               \vskip-\thickarrayrulewidth
             \fi
      \ifnum0=`{\fi}}
\makeatother

\newlength{\thickarrayrulewidth}
\graphicspath{{fig/}}

\newcommand{\R}{\mathbb{R}}

\newcommand{\N}{\mathbb{N}}

\newcommand{\Ord}{\mathrm{Ord}}
\newcommand{\Ext}{\mathrm{Ext}}
\newcommand{\Rel}{\mathrm{Rel}}
\newcommand{\Dg}{\mathrm{Dg}}

\newcommand{\hks}[1]{$\mathrm{hks}_{#1}$}

\newcommand{\LapW}{L_w}

\newtheorem{thm}{Theorem}[section]

\newtheorem{defin}[thm]{Definition}

\newcommand{\bottleneck}{\mathrm{d}_B}

\begin{document}

%

%

\title{
\textsc{PersLay}: A Neural Network Layer for Persistence Diagrams and New Graph Topological Signatures}

\author{Mathieu Carri\`ere, Fr\'ed\'eric Chazal,  Yuichi Ike, \\
Th\'eo Lacombe, Martin Royer, Yuhei Umeda} 





\maketitle 

\begin{abstract}
Persistence diagrams, the most common descriptors of Topological Data Analysis, encode topological properties of data and have already proved pivotal in many different applications of data science. However, since the metric space of persistence diagrams is not Hilbert, they end up being difficult inputs for most Machine Learning techniques. To address this concern, several vectorization methods have been put forward that embed persistence diagrams into either finite-dimensional Euclidean space or implicit infinite dimensional Hilbert space with kernels.

{In this work, we focus on persistence diagrams built on top of graphs. Relying on \emph{extended} persistence theory and the so-called heat kernel signature, we show how graphs can be encoded by (extended) persistence diagrams in a provably stable way. We then propose a general and versatile framework for learning vectorizations of persistence diagrams, which encompasses most of the vectorization techniques used in the literature. We finally showcase the experimental strength of our setup by achieving competitive scores on classification tasks on real-life graph datasets.}
\end{abstract}

\section{Introduction}
\label{sec:intro}

Topological Data Analysis (TDA) is a field of data science whose goal is to detect and encode topological features (such as connected components, loops, cavities...) that are present in datasets in order to improve inference and prediction.
Its main descriptor is the so-called {\em persistence diagram}, which takes the form of a set of points in the Euclidean plane $\R^2$, each point corresponding to a topological feature of the data, with its coordinates encoding the feature size.
This descriptor has been successfully used in many different applications of data science, such as signal analysis~\cite{Perea2015}, material science~\cite{Buchet2018}, cellular data~\cite{Camara2017}, or shape recognition~\cite{Li2014} to name a few.
This wide range of applications is mainly due to the fact that persistence diagrams encode information based on topology, and as such this information is very often complementary to the one retrieved by more classical descriptors.

However, the space of persistence diagrams heavily lacks structure: different persistence diagrams may have different number of points, and several basic operations are not well-defined, such as addition and scalar multiplication, which unfortunately dramatically impedes their use in machine learning applications.
To handle this issue, a lot of attention has been devoted to {\em vectorizations} of persistence diagrams through the construction of either {\em finite-dimensional embeddings}~\cite{Adams2017, Carriere2015, Chazal2015, Kalisnik2018}, i.e., embeddings turning persistence diagrams into vectors in Euclidean space $\R^d$, or {\em kernels}~\cite{Bubenik2015, Carriere2017, Kusano2016, Le2018, Reininghaus2015}, i.e., generalized scalar products that implicitly turn persistence diagrams into elements of infinite-dimensional Hilbert spaces. 

Even though these methods improved the use of persistence diagrams in machine learning tremendously, several issues remain. 
For instance, most of these vectorizations only have a few trainable parameters, which may prevent them from fitting well to specific applications. As a consequence, it may be very difficult to determine which vectorization is going to work best for a given task. Furthermore, kernel methods {(which are generally efficient in practice)} require to compute and store the kernel evaluations for each pair of persistence diagrams. Since all available kernels have a complexity that is at least linear, and often quadratic in the number of persistence diagram points for a single matrix entry computation, kernel methods quickly become very expensive in terms of running time and memory usage on large sets or for large diagrams.

In this work, we show how to use neural networks for handling persistence diagrams. Contrary to static vectorization methods proposed in the literature, we actually learn the vectorization with respect to the learning task that is being solved. Moreover, our framework is general enough so that most of the common vectorizations of the literature~\cite{Adams2017, Bubenik2015, Chazal2015} can be retrieved from our method by specifying its parameters accordingly.

\subsection{Our contributions}

{The contribution of this paper is two-fold.}

{First, we introduce in \cref{sec:hks_theory} a new family of topological signatures on graphs: the \emph{extended} persistence diagrams built from the \emph{Heat Kernel Signatures} (HKS) of the graph. These signatures depend on a diffusion parameter $t \geqslant 0$. Although HKS are well-known signatures, they have never been used in the context of persistent homology to encode topological information for graphs. We prove that the resulting diagrams are stable with respect to both the input graph and the parameter $t$. The use of extended persistence, by opposition to the commonly used \emph{ordinary} persistence, is introduced in order to handle ``essential'' components, see \cref{subsec:extended_persistence} below. To our knowledge, it is the first use of extended persistence in a machine learning context. We also experimentally showcase its strength over ordinary persistence on several applications.}

{Second, building on the recent introduction of \textit{Deep Sets} from \cite{Zaheer2017}, we apply and extend that framework for persistence diagrams, implementing {\textsc{PersLay}}: a simple, highly versatile, automatically differentiable layer for neural network architectures that can process topological information encoded in persistence diagrams computed from all sorts of datasets. Our framework encompasses most of the common vectorizations that exist in the literature, and we give a necessary and sufficient condition for the learned vectorization to be continuous, improving on the analysis of \cite{Hofer2019JMLR}. Using a large-scale dataset coming from dynamical systems which is commonly used in the TDA literature, we give in \cref{subsec:orbit} a proof-of-concept of the scalability and efficiency of this neural network approach over standard methods in TDA. The implementation of \textsc{PersLay} is publicly available\footnote{\url{https://github.com/MathieuCarriere/perslay}} as a plug-and-play Python package based on \texttt{tensorflow}, as well as a module of the \texttt{Gudhi}\footnote{\url{http://gudhi.gforge.inria.fr/python/latest/}} library.} 

{We finally combine these two contributions in Section \ref{sec:expe} by performing} real-graph classification application with benchmark datasets coming various fields of science, such as biology, chemistry and social sciences.

\begin{figure*}
	\center
	\includegraphics[width=0.9\textwidth]{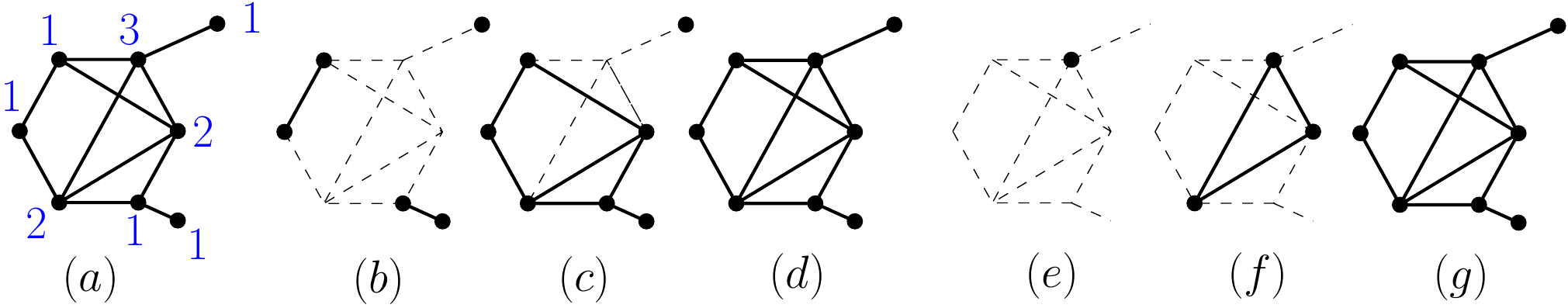}
	\caption{Illustration of sublevel and superlevel graphs. $(a)$ Input graph $(V, E)$ along with the values of a function $f : V \to \R$ (blue). $(b,c,d)$ Sublevel graphs for $\alpha=1, 2, 3$ respectively. $(e,f,g)$ Superlevel graphs for $\alpha=3, 2, 1$ respectively.}
	\label{fig:graph_sublevel}
\end{figure*}

\subsection{Related work}

Various techniques have been proposed to encode the topological information that is contained in the structure of a given graph, see for instance \cite{archambault2007topolayout, li2012effective, ferrara2012topological}. 
In this work, we focus on topological features computed with persistent homology (see \cref{subsec:background_ordinary_persistence} below).
This requires to define a real-valued function $f$ on the nodes of the graph. 
A first simple choice---made in \cite{hofer2017deep}---is to map each node to its degree. 
In another recent work \cite{Zhao2019}, authors proposed to use the Jaccard index and the Ricci curvature. 
In \cite{tran2018scale}, authors adopt a slightly different approach: given a graph with $N$ nodes indexed by $\{1\dots N\}$ and an integer parameter $\tau$, they compute an $N \times N$ matrix $(p_\tau(i|j))_{i,j}$ where $p_\tau(i|j)$ is the probability that a random walk starting at node $j$ ends at node $i$ after $\tau$ steps.
This matrix can be thought of as a finite metric space, i.e., a set of $N$ points embedded in $\R^N$, on which topological descriptors can also be computed~\cite{chazal2014persistence}. Here, $\tau$ acts as a scale parameter: small values of $\tau$ will encode local information while large values will catch large-scale features. Our approach, presented in \cref{sec:hks_theory}, shares the same scale-parametric idea, with the notable difference that we compute our topological descriptors on the graphs directly.

The first approach to feed a neural network architecture with a persistence diagram was presented in~\cite{hofer2017deep}.
It amounts to evaluating the points of the persistence diagram against one (or more) Gaussian distributions with parameters $(\mu,\sigma)$ that are learned during the training process (see Section \ref{sec:deepsetPD} for more details). Such a transformation is oblivious to any ordering of the diagram points, which is a suitable property, and is a particular case of \emph{permutation invariant} transformations. These general transformations are studied in \cite{Zaheer2017}, and used to define neural networks on sets. These networks are referred to as \emph{Deep Sets}. In particular, the authors in \cite{Zaheer2017} observed that any permutation invariant function $L$ defined on point clouds supported on $\R^p$ with exactly $n$ points can be written of the form:
\begin{equation} \label{eq:perm_inv}
	L(\{ x_1 \dots x_n \}) = \rho\left(\sum_{i=1}^n \phi(x_i) \right),
\end{equation}
for some $\phi : \R^p \to \R^q$ and $\rho : \R^q \to \R^q$. Obviously, the converse is true: any function defined by \eqref{eq:perm_inv} is a permutation invariant function.  Hofer et al.~make use of this idea in \cite{Hofer2019JMLR}, where they suggest three possible functions $\phi:\R^2\rightarrow\R$, which roughly correspond to Gaussian, spike and cone functions centered on the diagram points.
Our framework builds on the same idea, but substantially generalize theirs, as we are able to generate many more possible vectorizations and ways to combine them (for instance by using a maximum instead of a sum in Equation~\ref{eq:perm_inv}). We deepen the analysis by observing how common vectorizations of the TDA literature can be obtained again as specific instances of our architecture (\cref{sec:deepsetPD}). Moreover, we allow for more general weight functions that provide additional interpretability, as shown in Appendix, Section~\ref{appendix:complementary_expe_results}.


\subsection{Background on ordinary persistence}
\label{subsec:background_ordinary_persistence}

In this section, we briefly recall the basics of ordinary persistence theory. We refer the interested reader to~\cite{Cohen-Steiner2009, Edelsbrunner2010, Oudot2015} for a thorough description. 

Let $X$ be a topological space, and $f \colon X \to \R$ be a real-valued continuous function. The $\alpha$-\emph{sublevel set} of $X$ is then defined as: 
$X_\alpha=\{x \in X \,:\, f(x) \leq \alpha \}$.
Making $\alpha$ increase from $-\infty$ to $+\infty$ gives an increasing sequence of sublevel sets, called the {\em filtration} induced by $f$. 
It starts with the empty set and ends with the whole space $X$ (see $(b-d)$ in \cref{fig:graph_sublevel} for an illustration on a graph).
Ordinary persistence keeps track of the times of appearance and disappearance of topological features (connected components, loops, cavities, etc.) in this sequence.
For instance, one can store the value $\alpha_b$, called the \emph{birth time}, for which a new connected component appears in $X_{\alpha_b}$. 
This connected component eventually gets merged
with another one for some value $\alpha_d \geq \alpha_b$, which is stored as well and called the \emph{death time}. Moreover, one says that the component \emph{persists} on the corresponding interval $[\alpha_b,\alpha_d]$.
Similarly, we save the $[\alpha_b,\alpha_d]$ values of each loop, cavity, etc. that appears in a specific sublevel set $X_{\alpha_b}$ and disappears (get ``filled'') in $X_{\alpha_d}$. This family of intervals is called the barcode, or \emph{persistence diagram}, of $(X,f)$, and can be represented as a multiset of points (i.e., point cloud where points are counted with multiplicity) supported on $\R^2$ with coordinates $\{(\alpha_b,\alpha_d)\}$.

The space of persistence diagrams can be equipped with a parametrized metric $d_s$, $1 \leqslant s \leqslant \infty$, whose proper definition is not required in this work and is given in Appendix, \cref{appendix:hks} for the sake of completeness. In the particular case $s = \infty$, this metric will be refered to as the \emph{bottleneck} distance between persistence diagrams.

\section{Extended persistence diagrams}
\label{sec:hks_theory}

\begin{figure*}
	\centering
	\includegraphics[height=3cm]{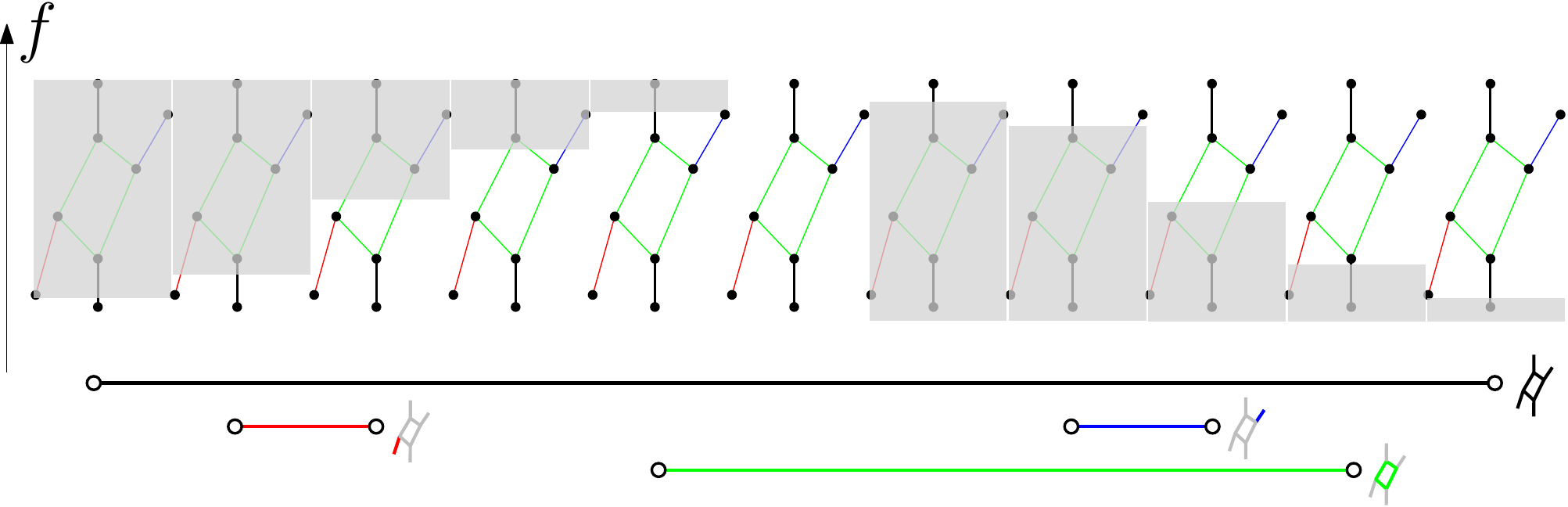}
	\hspace{1cm}
	\includegraphics[height=2.5cm]{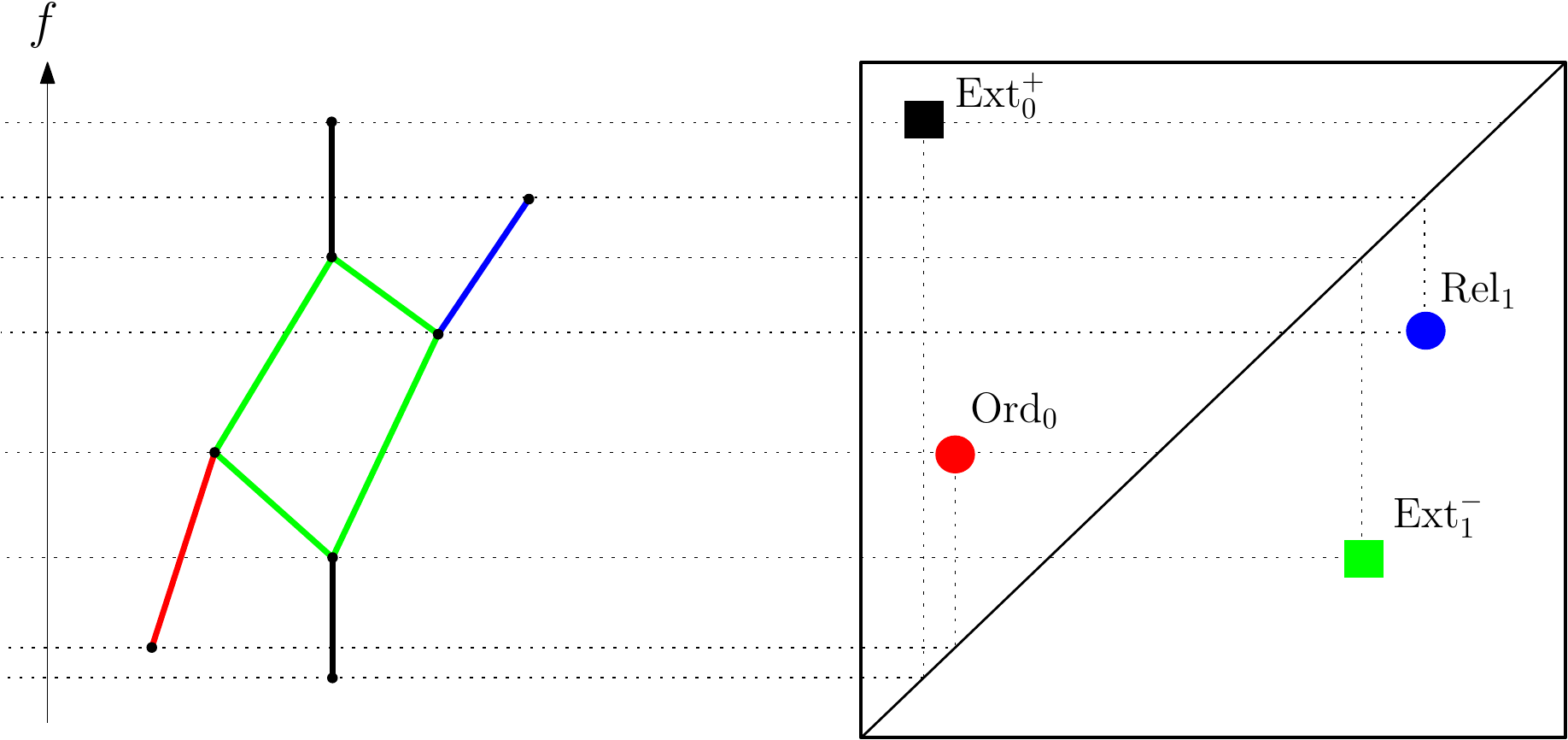}
	\caption{Extended persistence diagram computed on a graph: topological features of the graph are detected in the sequence of sublevel and superlevel graphs shown on the left of the figure.
	The corresponding intervals are displayed under the sequence: the black interval represents the connected component of the graph, the red one represents its downward branch, the blue one represents its upward branch, and the green one represents its loop. The extended persistence diagram given by the intervals is shown on the right.
	}
	\label{fig:expers}
\end{figure*}

\subsection{Extended persistence}
\label{subsec:extended_persistence}
In general ordinary persistence does not fully encode the topology of $X$. For instance, consider a graph $G = (V,E)$, with vertices $V$ and (non-oriented) edges $E$. Let $f \colon V\rightarrow\R$ be a function defined on its vertices, and consider the \emph{sublevel graphs} $G_\alpha=(V_\alpha,E_\alpha)$ where $\alpha\in\R$, $V_\alpha=\{v\in V\,:\,f(v)\leqslant \alpha\}$ , and $E_\alpha=\{(v_1,v_2)\in E\,:\,v_1,v_2\in V_\alpha\}$, see $(b-d)$ in Figure~\ref{fig:graph_sublevel}. 
In this sequence $(G_\alpha)_\alpha$, the loops persist forever since they never disappear from the sequence of sublevel graphs (they never get ``filled''), and the same applies for whole connected components of $G$. Moreover, branches pointing upwards (with respect to the orientation given by $f$, see Figure~\ref{fig:expers}) are missed (while those pointing downward are detected), since they do not create connected components when they appear in the sublevel graphs, making ordinary persistence unable to detect them.

To handle this issue, extended persistence refines the analysis by also looking at the so-called \emph{superlevel set} $X^\alpha=\{ x \in X \,:\, f(x) \geqslant \alpha \}$.
Similarly to ordinary persistence on sublevel sets, making $\alpha$ decrease from $+\infty$ to $-\infty$ produces a sequence of increasing subsets, for which structural changes can be recorded.

Although extended persistence can be defined for general metric spaces (see the references given above), we restrict ourselves to the case where $X=G$ is a graph.
The sequence of increasing superlevel graphs $G^{\alpha}$ is illustrated in Figure~\ref{fig:graph_sublevel} $(e-g)$.
In particular, death times can be defined for loops and whole connected components by picking the superlevel graphs for which the feature appears again, and using the corresponding $\alpha$ value as the death time for these features.
In this case, branches pointing upwards can be detected in this sequence of superlevel graphs, in the exact same way that downwards branches were in the sublevel graphs.
See Figure~\ref{fig:expers} for an illustration.

Finally, the family of intervals of the form $[\alpha_b,\alpha_d]$ is turned into a multiset of points in the
Euclidean plane $\R^2$ by using the interval endpoints as coordinates. This multiset is called the \emph{extended persistence diagram} of $f$ and is denoted by $\Dg(G,f) \subset \R^2$.

Since graphs have four types of topological features (see Figure~\ref{fig:expers}), namely upwards branches, downwards branches, loops and connected components, the corresponding points in extended persistence diagrams
can be of four different types. These types are denoted as $\Ord_0$, $\Rel_1$, $\Ext_0^+$ and $\Ext_1^-$ for downwards branches, upwards branches, connected components and loops respectively.

While it encodes more information than ordinary persistence, extended persistence ensures that points have finite coordinates.
In comparison, methods relying on ordinary persistence have to design specific tools to handle points with infinite coordinates~\cite{hofer2017deep, Hofer2019JMLR}, or simply ignore them~\cite{Carriere2017}, losing information in the process. Therefore extended persistence allows the use of generic architectures regardless of the homology dimension.
Empirical performances show substantial improvement over using ordinary persistence only (see Appendix, \cref{tab:complementary_expe}).
In practice, computing extended persistence diagrams can be efficiently done with the C++/Python \texttt{Gudhi} library~\cite{gudhi}. Persistence diagrams are usually compared with the so-called bottleneck distance $\bottleneck$---whose proper definition is not required for this work and is recalled in Appendix, Section~\ref{appendix:hks}. However, the resulting metric space is not Hilbert and as such, incorporating diagrams in a learning pipeline requires to design specific tools, which we do in \cref{sec:deepsetPD}.

We recall that extended persistence diagrams can be computed only after having defined a real-valued function on the nodes of the graphs. In the next section, {we define a family of such functions from the so-called Heat Kernel Signatures (HKS) for graphs, and show that these signatures enjoy stability properties. Moreover, Section~\ref{sec:expe} will further demonstrate that they lead to competitive results for graph classification.}

\subsection{Heat kernel signatures}
\label{subsec:hks}
HKS is an example of spectral family of signatures, that is, functions derived from the spectral decomposition of graph Laplacians, which provide informative features for graph analysis. We start this section with a few basic definitions. The adjacency matrix $A$ of a graph $G$ with vertex set $V = \{ v_1, \dots, v_n \}$ is the matrix $A := (\boldsymbol{1}_{(v_i,v_j) \in E})_{i,j}$. The degree matrix $D$ is the diagonal matrix defined by $D_{i,i} = \sum_j A_{i,j}$. The normalized graph Laplacian $\LapW = \LapW(G)$ is the linear operator acting on the space of functions defined on the vertices of $G$, and is represented by the matrix $\LapW = I - D^{-\frac{1}{2}} A D^{-\frac{1}{2}}$. It admits an orthonormal basis of eigenfunctions $\Psi = \{ \psi_1, \dots, \psi_n \}$ and its eigenvalues satisfy $0 \leqslant \lambda_1 \leqslant\dots \leqslant\lambda_n \leqslant 2$. As the orthonormal eigenbasis $\Psi$ is not uniquely defined, the eigenfunctions $\psi_i$ cannot be used as such to compare graphs. Instead we consider the \emph{Heat Kernel Signatures} (HKS):
\begin{defin}[\cite{hu2014stable, sun2009concise}]
Given a graph $G$ and $t \geq 0$, the \emph{Heat Kernel Signature} with diffusion parameter $t$ is the function $\mathrm{hks}_{G,t}$ defined on the vertices of $G$ by 
$\mathrm{hks}_{G,t} \colon v\mapsto \sum_{k=1}^n \exp(- t \lambda_k) \psi_k(v)^2$.
\end{defin}

The HKS have already been used as signatures to address graph matching problems \cite{hu2014stable} or to define spectral descriptors to compare graphs \cite{tsitsulin2018netlsd}. These signatures rely on the distributions of values taken by the HKS but not on their global topological structures, which are encoded in their extended persistence diagrams.
For the sake of concision, we denote by $\Dg(G,t)$ the extended persistence diagram obtained from a graph $G$ using the filtration induced by the HKS with diffusion parameter $t$, that is $\Dg(G, \mathrm{hks}_{G,t})$.
The following theorem shows these diagrams to be stable with respect to the bottleneck distance $\bottleneck$ between persistence diagrams. The proof can be found in Appendix, Section~\ref{appendix:hks}.

\begin{thm} \label{thm:stab-hks-G}
{\bf Stability w.r.t.~graph perturbations.} Let $t \geqslant 0$ and let $\LapW$ be the Laplacian matrix of a graph $G$ with $n$ vertices. Let $G'$ be another graph with $n$ vertices and Laplacian matrix
 $\tilde{L}_w = \LapW + W$. Then there exists a constant $C(G,t) > 0$ only depending on $t$ and the spectrum of $\LapW$ such that, for small enough $\| W \|_F$, where $\| \cdot \|_F$ denotes the Frobenius norm:
 \begin{align}
 	 \bottleneck (\Dg(G, t), \Dg(G', t)) \leqslant C(G,t) \| W \|_F.
 \end{align}
\end{thm}

\paragraph{On the influence of diffusion parameter $t$.}

Building an extended persistence diagram on top of a graph $G$ with the Heat Kernel Signatures requires to pick a specific value of $t$. In particular, understanding the influence and thus the choice of the diffusion parameter $t$ is an important question for statistical and learning applications. First, we state that the map $t \mapsto \Dg(G,t)$ is Lipschitz-continuous. The proof is found in Appendix, Section~\ref{appendix:hks}.
\begin{thm} \label{thm:stab-hks-t}
{\bf Stability w.r.t.~parameter $t$.} Let $G$ be a graph. The map $t \mapsto \Dg(G,t)$ is $2$-Lipschitz continuous, that is for $t,t' \in \R$,
\begin{equation}
	\bottleneck(\Dg(G,t), \Dg(G,t')) \leqslant 2|t - t'|
\end{equation}
\end{thm}
It follows from Theorem~\ref{thm:stab-hks-t} that persistence diagrams are robust to the choice of $t$. An empirical illustration is shown in Appendix, Figures \ref{fig:hks_evol} and \ref{fig:accuracy_wrt_t}.

\section{Neural network learning with \textsc{PersLay}}
\label{sec:deepsetPD}
In this section, we introduce \textsc{PersLay}: a general and versatile neural network layer for learning persistence diagram vectorizations.

\subsection{PersLay}
\label{subsec:perslay}
In order to define a layer for persistence diagrams, we modify the Deep Set architecture \cite{Zaheer2017} by defining and implementing a series of new permutation invariant layers, so as to be able to recover and generalize standard vectorization methods used in Topological Data Analysis. To that end we define our generic neural network layer for persistence diagrams, that we call \textsc{PersLay}, through the following equation:
\begin{equation}\label{eq:deepsetPD}
\mathrm{\textsc{PersLay}}(\Dg) :=  \texttt{op}\left(\{w(p)\cdot \phi(p)\}_{p\in\Dg}\right),
\end{equation}
where $\texttt{op}$ is any permutation invariant operation (such as minimum, maximum, sum, $k$th largest value...), $w:\R^2\rightarrow\R$ is a weight function for the persistence diagram points, and $\phi:\R^2\rightarrow\R^q$ is a representation function that we call \textit{point transformation}, mapping each point $(\alpha_b, \alpha_d)$ of a persistence diagram to a vector.

In practice, $w$ and $\phi$ are of the form $w_{\theta_1}, \phi_{\theta_2}$ where the gradients of $\theta_1 \mapsto w_{\theta_1}$ and $\theta_2 \mapsto \phi_{\theta_2}$ are known and implemented so that back-propagation can be performed, and the parameters $\theta_1,\theta_2$ can be optimized during the training process. We emphasize that any neural network architecture $\rho$ can be composed with \textsc{PersLay} to generate a neural network architecture for persistence diagrams.
Let us now introduce three point transformation functions
that we use and implement for parameter $\phi$ in Equation~(\ref{eq:deepsetPD}).
\begin{itemize}
	\item The \emph{triangle point transformation} $\phi_\Lambda:\R^2\rightarrow\R^q, p\mapsto \begin{bmatrix}
	\Lambda_p(t_1), \Lambda_p(t_2), \dots, \Lambda_p(t_q)
	\end{bmatrix}^T$ where the triangle function $\Lambda_p$ associated to a point $p=(x,y)\in\R^2$ is $\Lambda_p \colon t\mapsto \max \{0, y-|t-x|\}$, with $q\in\N$ and $t_1,\dots,t_q\in\R$. 

	\item The \emph{Gaussian point transformation} $\phi_\Gamma:\R^2\rightarrow\R^q, p\mapsto
	\begin{bmatrix}
	\Gamma_p(t_1), \Gamma_p(t_2), \dots, \Gamma_p(t_q)
	\end{bmatrix}^T$, where the Gaussian function $\Gamma_p$ associated to a point $p=(x,y)\in\R^2$ is $\Gamma_p \colon t\mapsto \exp\left(- \|p-t\|_2^2 / (2\sigma^2) \right)$ for a given $\sigma > 0$, $q\in\N$ and $t_1,\dots,t_q\in\R^2$. 

	\item The \emph{line point transformation} $\phi_L \colon \R^2\rightarrow\R^q, p\mapsto
	\begin{bmatrix}
	L_{\Delta_1}(p), L_{\Delta_2}(p), \dots, L_{\Delta_q}(p)
	\end{bmatrix}^T$, where the line function $L_\Delta$ associated to a line $\Delta$ with direction vector $e_\Delta\in\R^2$ and bias $b_\Delta\in\R$ is $L_\Delta:p\mapsto \langle p,e_\Delta \rangle + b_\Delta$, with $q\in\N$ and $\Delta_1,\dots,\Delta_q$ are $q$ lines in the plane.
\end{itemize}


Formulation \eqref{eq:deepsetPD} is very general: despite its simplicity, it allows to remarkably encode most of classical persistence diagram vectorizations with a very small set of point transformation functions $\phi$, allowing to consider the choice of $\phi$ as a hyperparameter of sort. Let us show how it connects to most of the popular vectorizations and kernel methods for persistence diagrams in the literature.

\begin{itemize}
\item Using:
$\phi=\phi_\Lambda$ with samples $t_1,\dots,t_q\in\R$,
$\texttt{op}=k$th largest value,
$w = 1$ (a constant weight function),
amounts to evaluating the $k$th \emph{persistence landscape}~\cite{Bubenik2015} on $t_1,\dots,t_q\in\R$.

\item Using 
$\phi=\phi_\Lambda$ with samples $t_1,\dots,t_q\in\R$,
$\texttt{op}=$ sum,
arbitrary weight function $w$,
amounts to evaluating the \emph{persistence silhouette} weighted by $w$~\cite{Chazal2015} on $t_1,\dots,t_q\in\R$.

\item Using
$\phi=\phi_\Gamma$ with samples $t_1,\dots,t_q\in\R^2$,
$\texttt{op}=$ sum,
arbitrary weight function $w$,
amounts to evaluating the \emph{persistence surface} weighted by $w$~\cite{Adams2017} on $t_1,\dots,t_q\in\R^2$. Moreover, characterizing points of persistence diagrams with
Gaussian functions is also the approach advocated in several kernel methods for persistence diagrams~\cite{Kusano2016, Le2018, Reininghaus2015}.

\item Using 
$\phi=\phi_{\tilde \Gamma}$ where $\tilde\Gamma$ is a modification of the Gaussian point transformation defined with:
$\tilde \Gamma_p = \Gamma_{\tilde p}$ for any $p=(x,y)\in\R^2$, where $\tilde p = p$ if $y\leq\nu$ for some $\nu>0$, and $\left(x,\nu+\log\left(\frac{y}{\nu}\right)\right)$ otherwise,
$\texttt{op}=$ sum,
weight function $w = 1$,
is the approach presented in~\cite{hofer2017deep}.

\item Using 
$\phi=\phi_L$ with lines $\Delta_1,\dots,\Delta_q\in\R^2$,
$\texttt{op}=k$th largest value,
weight function $w = 1$,
is similar to the approach advocated in~\cite{Carriere2017}, where the sorted projections of the points onto the lines are then compared with the $\|\cdot\|_1$ norm
and exponentiated to build the so-called Sliced Wasserstein kernel for persistence diagrams.

\end{itemize}


\textbf{Stability of \textsc{PersLay}.} The question of the continuity and stability of persistence diagram vectorizations is of importance for TDA practitioners. In \cite[Remark 8]{Hofer2019JMLR}, authors observed that the operation defined in \eqref{eq:deepsetPD}---with \texttt{op}=\texttt{sum}---is not continuous in general (with respect to the common persistence diagram metrics). Actually, \cite[Prop.~5.1]{divol2019understanding} showed that for all $s \geqslant 1$ the map $(\Dg \mapsto \sum_{p \in \Dg} \phi(p))$ is continuous with respect to the metric $d_s$ if and only if $\phi$ is of the form $\phi(p) = \varphi(p)\|p - \Delta\|^s$, where $\|p - \Delta\|$ denotes the distance from a point $p \in \R^2$ to the diagonal $\Delta = \{(x,x),\ x \in \R\}$ and $\varphi$ is a continuous and bounded function. Furthermore, when $s=1$ and $\varphi$ is $1$-Lipschitz continuous, one can show that the map is actually stable (\cite[Thm.~12]{Hofer2019JMLR}, \cite[Prop.~5.2]{divol2019understanding}), in the following sense:
\[ \left\| \sum_{p \in \Dg_1} \phi(p) - \sum_{p' \in \Dg_2} \phi(p') \right\|_\infty \leqslant d_1(\Dg_1,\Dg_2). \]
In particular, this means that requiring continuity for the learned vectorization, as done in \cite{Hofer2019JMLR}, implies constraining the weight function to take small values for points close to the diagonal. However, in general there is no specific reason to consider that points close to the diagonal are less important than others, given a learning task. 

\subsection{A proof of concept: classification on large scale dynamical system dataset}
\label{subsec:orbit}

Our first application is on a synthetic dataset used as a benchmark in Topological Data Analysis \cite{Adams2017, Carriere2017, Le2018}. It consists in sequences of points generated by different dynamical systems, see \cite{hertzsch2007dna}. Given some initial position $(x_0, y_0) \in [0,1]^2$ and a parameter $r > 0$, we generate a point cloud $(x_n,y_n)_{n = 1, \dots, N}$ following:
\begin{align}
	\begin{cases}
		x_{n+1} = x_n + r y_n (1 - y_n) 		&\text{ mod } 1 \\
		y_{n+1} = y_n + r x_{n+1} (1 - x_{n+1}) &\text{ mod } 1		
	\end{cases}
\end{align}
The orbits of this dynamical system heavily depend on parameter $r$. More precisely, for some values of $r$, voids might form in these orbits (see Appendix, \cref{fig:orbits}), and as such, persistence diagrams are likely to perform well at attempting to classify orbits with respect to the value of $r$ generating them. As in previous works \cite{Adams2017, Carriere2017, Le2018}, we use the five different parameters $r = 2.5, 3.5, 4.0, 4.1$ and $4.3$ to simulate the different classes of orbits, with random initialization of $(x_0, y_0)$ and $N = 1,000$ points in each simulated orbit. These point clouds are then turned into persistence diagrams using a standard geometric filtration \cite{chazal2014persistence}, called the \texttt{AlphaComplex} filtration\footnote{\url{http://gudhi.gforge.inria.fr/python/latest/alpha_complex_ref.html}} in dimensions $0$ and $1$. We generate two datasets:
The first is 
$\texttt{ORBIT5K}$, where for each value of $r$, we generate $1,000$ orbits, ending up with a dataset of $5,000$ point clouds. This dataset is the same as the one used in \cite{Le2018}.
The second is
$\texttt{ORBIT100K}$, which contains $20,000$ orbits per class, resulting in a dataset of $100,000$ point clouds---a scale that kernel methods cannot handle. This dataset aims to show the edge of our neural-network based approach over kernels methods when dealing with very large datasets of large diagrams, 
since all the previous works dealing with this data \cite{Adams2017, Carriere2017, Le2018} use kernel methods. 

Results are displayed in \cref{tab:results_orbits}. 
Not only do we improve on previous results for $\texttt{ORBIT5K}$, we also show with $\texttt{ORBIT100K}$ that classification accuracy is further increased as more observations are made available. 
For consistency we use the same accuracy metric as \cite{Le2018}, that is, we split observations in 70\%-30\% training-test sets and report the average test accuracy over $100$ runs. 
The parameters used are summarized in Appendix, Section~\ref{appendix:complementary_expe_results}.

\begin{table*}
\begin{center}

\resizebox{.8\textwidth}{!}{%
\begin{tabular}{|l|cccc||c|}
\cline{1-6}
	Dataset   & PSS-K            & PWG-K           & SW-K           & PF-K           & \textsc{PersLay} \\
\cline{1-6}
	$\texttt{ORBIT5K}$   & 72.38($\pm$2.4)  & 76.63($\pm$0.7) & 83.6($\pm$0.9) & 85.9($\pm$0.8) & \textbf{87.7($\pm$1.0)} \\
	$\texttt{ORBIT100K}$ & ---              & ---             & ---            & ---            & \textbf{89.2($\pm$0.3)} \\
\cline{1-6}
\end{tabular}}
\end{center}
\caption[Caption for result]{Performance table. PSS-K, PWG-K, SW-K, PF-K stand for \emph{Persistence Scale Space Kernel} \cite{Reininghaus2015}, \emph{Persistence Weighted Gaussian Kernel} \cite{Kusano2016}, \emph{Sliced Wasserstein Kernel} \cite{Carriere2017} and \emph{Persistence Fisher Kernel} \cite{Le2018} respectively. We report the scores given in \cite{Le2018} for competitors on $\texttt{ORBIT5K}$, and the one we obtained using \textsc{PersLay} for both the $\texttt{ORBIT5K}$ and \texttt{ORBIT100K} datasets.
}
\label{tab:results_orbits}
\vskip-0.5cm
\end{table*}

\section{Application to graph classification}
\label{sec:expe}
\begin{figure}
	\centering
	\includegraphics[width=0.86\columnwidth]{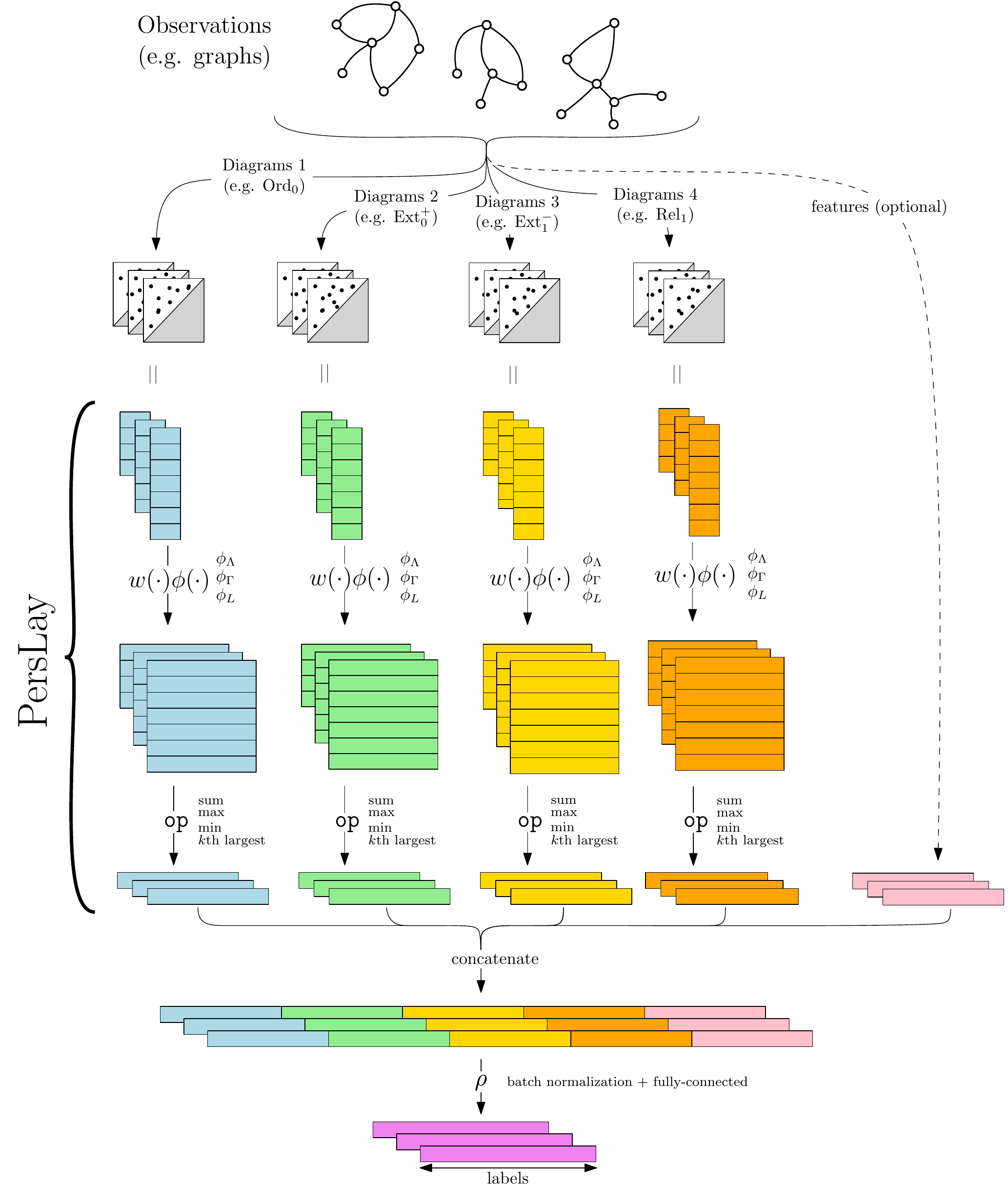}
	\caption{Network architecture illustrated in the case of our graph classification experiments (\cref{sec:expe}). Each graph is encoded as a set of persistence diagrams, then processed by an independent instance of \textsc{PersLay}. Each instance embeds diagrams in some vector space using two functions $w,\phi$ that are optimized during training and a fixed permutation-invariant operator \texttt{op}.
	}
	\label{fig:networkarchi}
\end{figure}

In order to truly showcase the contribution of \textsc{PersLay}, we use a very simple network architecture, namely a two-layer network. The first layer is \textsc{PersLay}, which processes persistence diagrams. The resulting vector is normalized and fed to the second and final layer, a fully-connected layer whose output is used for predictions. See \cref{fig:networkarchi} for an illustration. We emphasize that this simplistic two-layer architecture is designed so as to produce knowledge and understanding (see Appendix, Section~\ref{appendix:complementary_expe_results}), rather than achieving the best possible performances.

\paragraph{Choice of hyperparameters.}

In our experiments, we set $w : p \mapsto w_{i,j} \mathbf{1}_{p \in C_{i,j}}$, where $C_{i,j}$ denote the $(i,j)$-th cell in a $N \times N$ grid discretization of the unit square, and all $(w_{i,j})_{1 \leqslant i,j \leqslant N}$ are trainable parameters.
$N$ is typically set to $10$ or $20$. For aggregation operator \texttt{op} we use the sum. Further details are given in Appendix, see \cref{tab:expe_settings} for reporting of the chosen hyper-parameters and \cref{tab:influence_grid_size} for a study of the influence of the grid size or the choice of $\phi$.

Point transformations $\phi$ are chosen among the three choices $\{ \phi_\Lambda, \phi_\Gamma, \phi_L \}$ introduced in \cref{subsec:perslay}. Empirically no representation is uniformly better than the others.
The choice of the best point transformation $\phi$ for a given task could also be selected through a cross-validation scheme, or by learning a linear interpolation between these point transformations: by setting $\phi = \alpha_\Lambda \phi_\Lambda + \alpha_\Gamma \phi_\Gamma + \alpha_L \phi_L$, where $\alpha_\Lambda, \alpha_\Gamma, \alpha_L$ are trainable non-negative weights that sum to $1$. Thorough exploration of these alternatives is left for future work.

As mentioned in \cref{sec:hks_theory}, the diagrams we produce are stable with respect to the choice of the HKS diffusion parameter $t$ (Thm.~\ref{thm:stab-hks-G} and \ref{thm:stab-hks-t}).
As such, we generally use $t=0.1$ and $t=10$ in our experiments. 
We also refer to Appendix where \cref{fig:hks_evol} illustrates the evolution of a persistence diagram w.r.t.~$t$ and \cref{fig:accuracy_wrt_t} provides the classification accuracy through varying values of $t$.
In practice, it is thus sufficient to sample few values of $t$ using a log-scale, as suggested for example in \cite[\S 5]{sun2009concise}.\\
A subsequent natural question is: given a learning task, can $t$ itself be optimized? The question of optimizing over a family of filtrations induced by parametric functions $\{f_\theta\}_{\theta\in\Theta}$ the map $\theta \mapsto \Dg(G,f_\theta)$ has been studied both theoretically and practically in very recent works \cite{bruel2019topology, leygonie2019framework}. 
Hence, we also apply this approach for the filtrations induced by the HKS, optimizing the parameter $t$ during the learning process. Note that the running time of the experiments is greatly increased since one has to recompute all persistence diagrams for each epoch, that is, each time $t$ is updated. Moreover, we noticed after preliminary numerical investigations (see Appendix, Section~\ref{appendix:complementary_expe_results}) that classification accuracies were not improved by a large margin and remained comparable with results obtained without optimizing $t$, so we did not include this optimization step in our results.

\cref{tab:expe_settings} 
gives a detailed summary report of the different hyper-parameters chosen for each experiment.


\paragraph{Experimental settings.} We are now ready to evaluate our architecture on a series of different graph datasets commonly used as a baseline in graph classification problems. 
\texttt{REDDIT5K, REDDIT12K, COLLAB} (from \cite{deepgraphkernels}) \texttt{IMDB-B, IMDB-M} (from \cite{tran2018scale}) are composed of social graphs.
\texttt{COX2, DHFR, MUTAG, PROTEINS, NCI1, NCI109} are graphs coming from medical or biological frameworks (also from \cite{tran2018scale}).  
A quantitative summary of these datasets is found in Appendix, \cref{tab:data_desc}.

We compare performances with five other top graph classification methods.
Scale-variant topo \cite{tran2018scale} leverages a kernel for ordinary persistence diagrams computed on point cloud used to encode the graphs.
RetGK \cite{zhang2018retgk} is a kernel method for graphs that leverages eventual \emph{attributes} on the graph vertices and edges. 
FGSD \cite{verma2017hunt} is a finite-dimensional graph embedding that does not leverage attributes. 
Finally, GCNN \cite{xinyi2018capsule} and GIN \cite{Xu2018} are two graph neural network approaches that reach top-tier results. 
One could also compare our results on the \texttt{REDDIT} datasets to the ones of \cite{hofer2017deep}, where authors also use persistence diagrams to feed a network (using as first channel a particular case of \textsc{PersLay}, see \cref{sec:deepsetPD}), achieving 54.5\% and 44.5\% of accuracy on \texttt{REDDIT5K} and \texttt{REDDIT12K} respectively.

Topological features were extracted using the graph signatures introduced in \cref{sec:hks_theory}. We combine these features with more traditional graph features formed by the eigenvalues of the normalized graph Laplacian along with the deciles of the computed HKS (right-side channel in Figure~\ref{fig:networkarchi}).
In order to evaluate the impact of the topological features in this learning process, one will refer to the ablation study in Appendix, \cref{tab:complementary_expe}.

For each dataset, we perform 10 ten-fold evaluations and report the average and best ten-fold results. For a ten-fold evaluation, we split the data in 10 equally-sized folds, and record the classification accuracy obtained on the $i$-th fold (test) after training on the $9$ remaining others. Here, $i$ is cycled from 1 to 10. The mean of those 10 ten-fold experiments is naturally more robust for evaluation purposes, and we report it in the column ``\textsc{PersLay} - Mean''. This is consistent with the evaluation procedure from \cite{zhang2018retgk}.
Simultaneously, we also report the best single 10-fold accuracy obtained, reported in the column ``\textsc{PersLay} - Max'', which is comparable to the results reported by all the other competitors.

In most cases, our approach is comparable with state-of-the-art results, despite using a very simple neural network architecture.
Interestingly, both topology-based methods (SV and \textsc{PersLay}) have mediocre performances on the \texttt{NCI} datasets, suggesting that topology is not discriminative for these datasets.
Additional experimental results, including ablation studies and variations of hyper-parameters (weight grid size $N$, diffusion parameter $t$) are provided in~Appendix, Section~\ref{appendix:complementary_expe_results}.


\begin{table}
\begin{center}

\resizebox{\columnwidth}{!}{%
\begin{tabular}{|l|ccccc||cc|}
\cline{1-8}
	Dataset   & SV\footnotemark[1] & RetGK$^{*}$ \footnotemark[2] & FGSD \footnotemark[3] & GCNN \footnotemark[4] & GIN \footnotemark[5] & \multicolumn{2}{c|}{\textsc{PersLay}} \\
	          &                              &                               &                       &                       &                      & Mean            &  Max                \\
\cline{1-8}
	$\texttt{REDDIT5K}$        & ---  & 56.1 & 47.8 & 52.9 & 57.0                                  & 55.6 & 56.5 \\  
	$\texttt{REDDIT12K}$       & ---  & 48.7 & ---  & 46.6 & ---                  & 47.7 & 49.1 \\
	$\texttt{COLLAB}$          & ---  & 81.0 & 80.0 & 79.6 & 80.1               & 76.4 & 78.0\\
	$\texttt{IMDB-B}$          & 72.9 & 71.9 & 73.6 & 73.1 & 74.3                 & 71.2 & 72.6 \\
	$\texttt{IMDB-M}$          & 50.3 & 47.7 & 52.4 & 50.3 & 52.1                 & 48.8 & 52.2 \\
	$\texttt{COX2}^*$     & 78.4 & 80.1 & --- & ---  & ---                                    & 80.9 & 81.6 \\
	$\texttt{DHFR}^*$     & 78.4 & 81.5 & ---  & --- & ---                                    & 80.3 & 80.9 \\
	$\texttt{MUTAG}^*$    & 88.3 & 90.3 & 92.1 & 86.7 & 89.0      & 89.8 & 91.5 \\
	$\texttt{PROTEINS}^*$ & 72.6 & 75.8 & 73.4 & 76.3 & 75.9                 & 74.8 & 75.9 \\
	$\texttt{NCI1}^*$     & 71.6 & 84.5 & 79.8 & 78.4 & 82.7        & 73.5 & 74.0 \\
	$\texttt{NCI109}^*$   & 70.5 & ---  & 78.8 & ---  & ---         & 69.5 & 70.1 \\
\cline{1-8}
\end{tabular}}
\end{center}
\caption[Caption for result]{Classification accuracy over benchmark graph datasets. 
Our results (PersLay, right hand side) are recorded from ten runs of a 10-fold classification evaluation (see \cref{sec:expe} for details).
``Mean'' is consistent with \cite{zhang2018retgk}\footnotemark[2], while ``Max'' should be compared to \cite{tran2018scale}\footnotemark[1], \cite{verma2017hunt}\footnotemark[3], \cite{xinyi2018capsule}\footnotemark[4] and \cite{Xu2018}\footnotemark[5], as it corresponds to the mean accuracy over a \emph{single 10-fold}.
The \textbf{*} indicates datasets that contain attributes (labels) on graph nodes and symmetrically the methods that leverage such attributes for classification purposes.
}
\label{tab:tableau_res}
\end{table}

\section{Conclusion}

In this article, we introduced a new family of topological signatures on graphs, that are both stable and well-formed for learning purposes. In parallel we defined a powerful and versatile neural network layer to process persistence diagrams called \textsc{PersLay}, which generalizes most of the techniques used to vectorize persistence diagrams that can be found in the literature---while optimizing them task-wise.

We showcase the efficiency of our approach by achieving state-of-the-art results on synthetic orbit classification coming from dynamical systems and being competitive on several graph classification problems from real-life data, while working at larger scales than kernel methods developed for persistence diagrams and remaining simpler than most of its neural network competitors. We believe that \textsc{PersLay} has the potential to become a central tool to incorporate topological descriptors in a wide variety of complex machine learning tasks based on neural networks.

Our code is freely available publicly at \texttt{https://github.com/MathieuCarriere/perslay} and is part of the \texttt{Gudhi}\footnote{\url{http://gudhi.gforge.inria.fr/python/latest/}} library.

\clearpage

\bibliographystyle{alpha}
\bibliography{biblio}

\clearpage

\appendix

\section{Proofs of stability theorems} 
\label{appendix:hks}

\paragraph{Definition of diagram distances.}
Recall (see \cref{subsec:background_ordinary_persistence}) that persistence diagrams are generally represented as multisets of points (i.e.~points counted with multiplicity) supported on the upper half plane $\Omega = \{(b,d) \in \R^2, d > b\}$. Let $\mu = \{x_1, \dots, x_n\}$ and $\nu = \{y_1, \dots, y_m\}$ be two such diagrams and $s \geq 1$ be a parameter. Note in particular that $n \neq m$ in general. Let $\Delta = \{(t,t), t \in \R\}$ denote the diagonal, and let $\Pi(\mu,\nu)$ denote the set of all bijections between $\mu \cup \Delta$ and $\nu \cup \Delta$. Then, the $s$-diagram distance between $\mu$ and $\nu$ is defined as:
\begin{equation}
d_s(\mu,\nu) = \inf_{\pi \in \Pi(\mu,\nu)} \left( \sum_{x \in \mu \cup \Delta} \|x - s(x)\|^p \right)^{\frac{1}{p}}.
\end{equation}
In particular, if $s = \infty$, we recover the bottleneck distance defined as:
\begin{equation}
\bottleneck(\mu,\nu) = \inf_{\pi \in \Pi(\mu,\nu)} \sup_{x \in \mu \cup \Delta} \|x - s(x)\|.
\end{equation}

\paragraph{Proof of Theorem \ref{thm:stab-hks-G}}
The proof directly follows from the following two theorems. 
This first one, proved in \cite{hu2014stable}, is a consequence of classical arguments from matrix perturbation theory. 

\begin{thm}[\cite{hu2014stable}, Theorem 1]
	Let $t \geq 0$ and let $\LapW$ be the Laplacian matrix of a graph $G$ with $n$ vertices. 
	Let $\lambda_1 < \cdots < \lambda_k$, $k \leq n$ be the distinct eigenvalues of $\LapW$ and denote by $\delta > 0$ the smallest distance between two distinct eigenvalues: $\delta = \min_{j = 1, \cdots, k-1} |\lambda_{j+1} - \lambda_j|$. 
	Let $G'$ be another graph with $n$ vertices and Laplacian matrix $\tilde{L}_w = \LapW + W$ with $\|W \| < \delta$, where $\| W \|$ denotes the Frobenius norm of $W$. Then, 
	if $k=n$, there exists a constant $C_0(G,t) >0$ such that for any vertex $v \in G$,
	\[
	| \mathrm{hks}_{G,t}(v) - \mathrm{hks}_{G',t}(v) | \leqslant  C_0(G,t) \| W \|; 
	\]
	if $k < n$, there exists two constants $C_1(G,t), C_2(G,t) >0$ such that for any vertex $v \in G$,
	\begin{align*}
		| \mathrm{hks}_{G,t}(v) - &\mathrm{hks}_{G',t}(v) | \leqslant \\ &C_1(G,t) \frac{\| W \|}{\delta - \| W \|} + C_2(G,t) \| W \|
	\end{align*}
\end{thm}  

In particular, if $\| W \| < \frac{\delta}{2}$, there exists a constant $C(G,t) > 0$---notice that $\delta$ also depends on $G$---such that in the two above cases, 
\[
| \mathrm{hks}_{G,t}(v) - \mathrm{hks}_{G',t}(v) | \leqslant C(G,t) \| W \|. 
\]
Theorem \ref{thm:stab-hks-G} then immediately follows from the second following theorem, which is a special case of general stability results for persistence diagrams. 

\begin{thm}[\cite{Chazal2016, Cohen-Steiner2009}]\label{th:dgstab}
	Let $G=(V,E)$ be a graph and $f,g:V\rightarrow\R$ be two functions defined on its vertices. Then:
	\begin{equation}
	\bottleneck(\Dg(G,f),\Dg(G,g)) \leqslant \|f-g\|_\infty,
	\end{equation}
	where $\bottleneck$ stands for the so-called {\em bottleneck distance} between persistence diagrams and 
	$\|f-g\|_\infty = \sup_{v \in G} |f(v) - g(v)|$. Moreover, this inequality is also satisfied for each of 
	the subtypes $\Ord_0,\Rel_1,\Ext_0^+$ and $\Ext_1^-$ individually.
\end{thm}

\paragraph{Proof of Theorem \ref{thm:stab-hks-t}}
Fix a graph $G = (V,E)$. With the same notations as in \cref{subsec:hks}, recall that the eigenvalues of the normalized graph Laplacian satisfy $0 \leq \lambda_1 \leq \dots \leq \lambda_n \leq 2$, and the corresponding eigenvectors $\{\psi_1, \dots, \psi_n\}$ define an orthonormal family. In particular, $t \mapsto \exp(-t\lambda_k)$ is $2$-Lipschitz continuous for $t > 0$. Let $t,t'$ be two positive diffusion parameters. We have, for any $v \in V$:
\begin{align*}
	  \Big| \sum_{k=1}^n & (\exp(-t \lambda_k) - \exp(-t' \lambda_k))\psi_k(v)^2 \Big| \\ 
          & \leqslant 2 \cdot |t'-t| \underbrace{\sum_{k=1}^n \psi_k(v)^2}_{=1}.
\end{align*}
Thus in particular,
\[ \sup_{v \in V} |\mathrm{hks}_{G,t}(v) - \mathrm{hks}_{G,t'}(v)| \leqslant 2 |t - t'|. \]
As in the previous proof, we conclude using the stability of persistence diagrams w.r.t. the bottleneck distance (see Thm.~\ref{th:dgstab}).

%


\section{Datasets description}

\begin{figure}[ht]
	\includegraphics[width=\columnwidth]{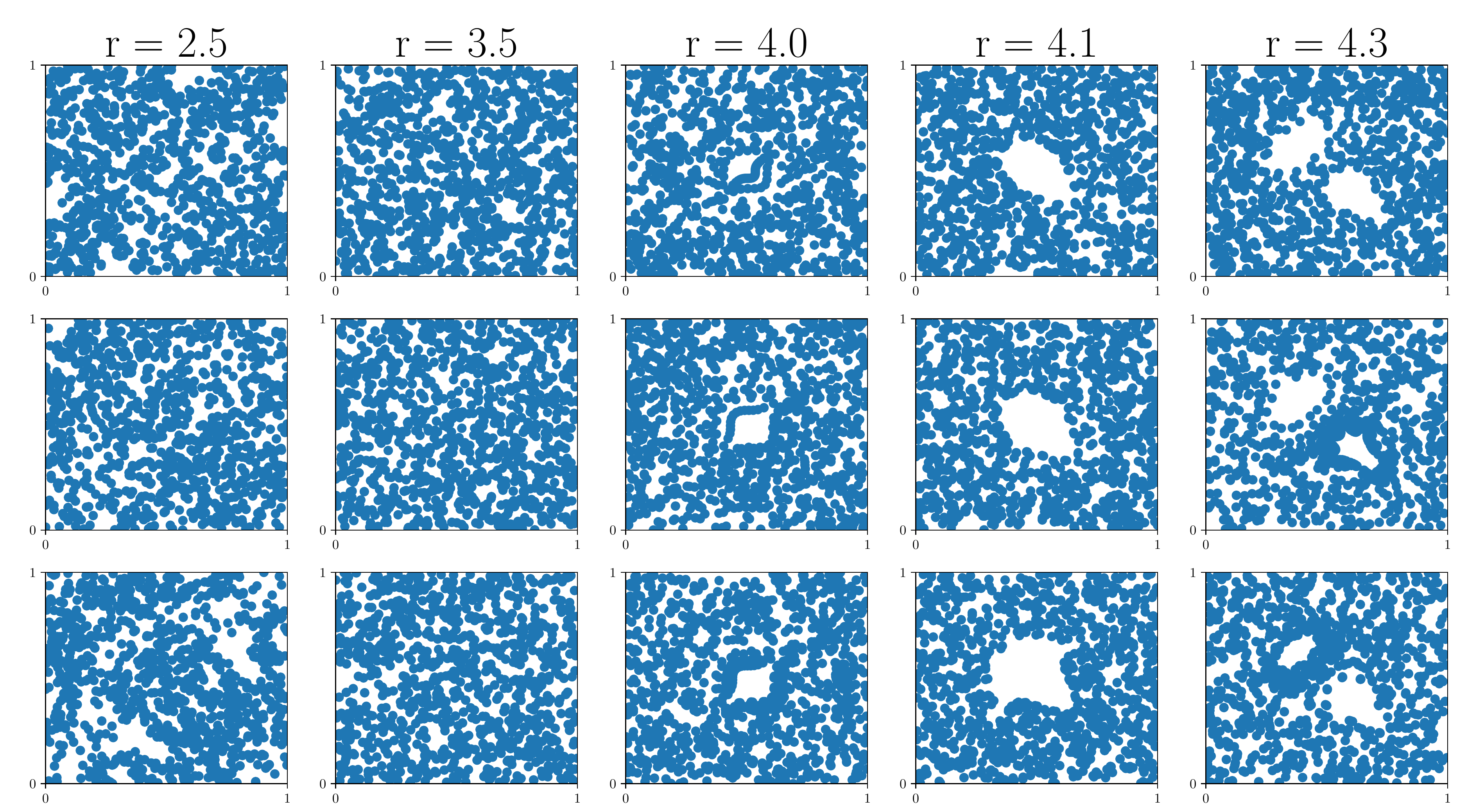}
	\caption{Some example of orbits generated by the different choices of $r$ (three simulations are represented for the different values of $r$).}
	\label{fig:orbits}
\end{figure}

Tables~\ref{tab:orbits_desc} and \ref{tab:data_desc} summarize key information of each dataset for both our experiments. We also provide in \cref{fig:orbits} an illustration of the orbits we generated in \cref{subsec:orbit}.
\begin{table*}
\begin{center}
\resizebox{.7\textwidth}{!}{%
\begin{tabular}{|l|c|c|c|}
\cline{1-4}
	Dataset & Nb of orbit observed & Number of classes & Number of points per orbit \\
\cline{1-4}
	$\texttt{ORBIT5K}$     & 5,000   & 5   & 1,000 \\
	$\texttt{ORBIT100K}$   & 100,000 & 5   & 1,000 \\
\cline{1-4}
\end{tabular}}
\end{center}
\caption{Description of the two orbits dataset we generated. The five classes correspond to the five parameter choices for $r \in \{ 2.5, 3.5, 4.0, 4.1, 4.3 \}$. In both $\texttt{ORBIT5K}$ and $\texttt{ORBIT100K}$, classes are balanced.}
\label{tab:orbits_desc}
\end{table*}

\begin{table*}
\begin{center}
\resizebox{.7\textwidth}{!}{%
\begin{tabular}{|l|c|c|cc|cc|}
\cline{1-7}
	Dataset & Nb graphs & Nb classes & Av.~nodes & Av.~Edges & Av.~$\beta_0$ & Av.~$\beta_1$ \\
\cline{1-7}
	$\texttt{REDDIT5K}$     & 5,000  & 5   & 508.5 & 594.9 & 3.71 & 90.1   \\
	$\texttt{REDDIT12K}$    & 12,000 & 11  & 391.4 & 456.9 & 2.8 & 68.29   \\
	$\texttt{COLLAB}$       & 5,000  & 3   & 74.5  & 2457.5& 1.0 & 2383.7  \\
	$\texttt{IMDB-B}$       & 1,000  & 2   & 19.77 & 96.53 & 1.0 & 77.76   \\
	$\texttt{IMDB-M}$   & 1,500  & 3   & 13.00 & 65.94 & 1.0 & 53.93   \\
	$\texttt{COX2}$         & 467   & 2   & 41.22 & 43.45 & 1.0 & 3.22  \\
	$\texttt{DHFR}$         & 756   & 2   & 42.43 & 44.54 & 1.0 & 3.12  \\
	$\texttt{MUTAG}$        & 188   & 2   & 17.93 & 19.79 & 1.0 & 2.86  \\
	$\texttt{PROTEINS}$     & 1,113  & 2   & 39.06 & 72.82 & 1.08 & 34.84\\
	$\texttt{NCI1}$         & 4,110  & 2   & 29.87 & 32.30 & 1.19 & 3.62 \\
	$\texttt{NCI109}$       & 4,127  & 2   & 29.68 & 32.13 & 1.20 & 3.64 \\
\cline{1-7}
\end{tabular}}
\end{center}
\caption{Datasets description. $\beta_0$ (resp.~$\beta_1$) stands for the $0$th-Betti-number (resp.~$1$st), that is the number of connected components (resp.~cycles) in a graph. 
In particular, an average $\beta_0 = 1.0$ means that all graph in the dataset are connected, and in this case $\beta_1 = \#\{\text{edges}\} - \#\{\text{nodes}\}$. }
\label{tab:data_desc}
\end{table*}

\section{Complementary experimental results}
\label{appendix:complementary_expe_results}
\subsection{Weight learning}
\label{appendix:weight}
\begin{figure*}[ht!]
\centering
  \includegraphics[height=3cm]{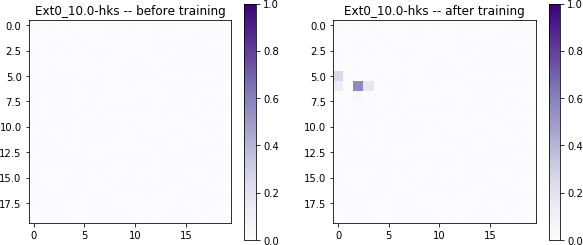}\ \ \ 
\includegraphics[height=3cm]{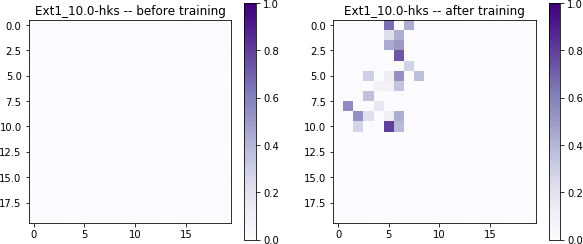}\ \ \ 
\includegraphics[height=3cm]{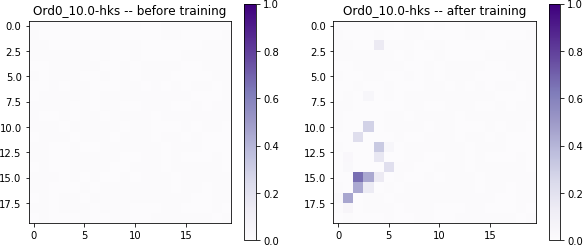}\ \ \ 
\includegraphics[height=3cm]{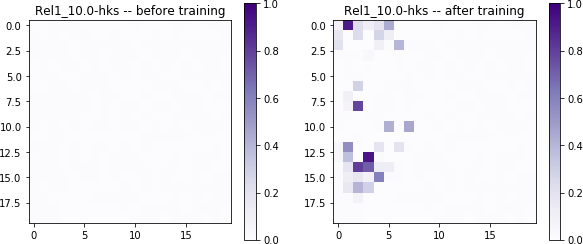}
  \caption{Weight function $w$ when chosen to be a grid with size $20 \times 20$ before and after training (\texttt{MUTAG} dataset). Here, Ord0, Rel1, Ext0, and Ext1 denote the extended diagrams corresponding to downwards branches, upwards branches, connected components and loops respectively (cf \cref{subsec:extended_persistence}).}
  \label{fig:weight_evol}
\end{figure*}

\cref{fig:weight_evol} provides an illustration of the weight grid $w$ learned after training on the \texttt{MUTAG} dataset. 
Roughly speaking, activated cells highlight the areas of the plane where the presence of points was discriminating in the classification process.
These learned grids thus emphasize the points of the persistence diagrams that matter w.r.t. learning task.

\subsection{Selection of HKS diffusion parameter}

\begin{figure}
	\begin{center}
		\includegraphics[width=0.5\columnwidth]{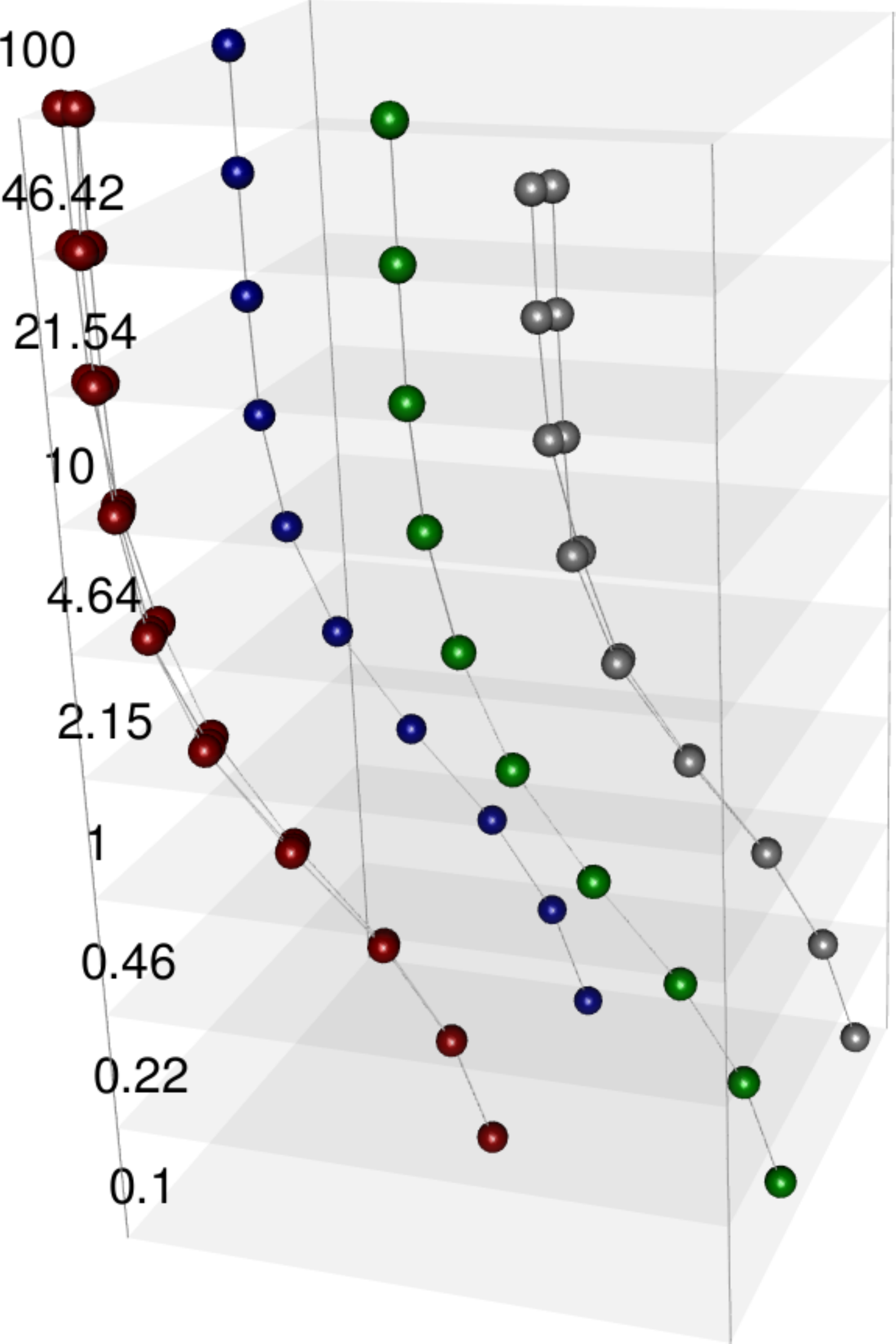}
	\end{center}
	\caption{Evolution of $t \mapsto \Dg(G,t))$ for one graph from the \texttt{MUTAG} dataset ($t \in [0.1,100]$, $t$ in log-scale).}
	\label{fig:hks_evol}
\end{figure}

As stated in Theorem \ref{thm:stab-hks-t}, for a fixed graph $G$, the function $t \mapsto \Dg(G, t)$ is $2$-Lipschitz continuous with respect to the bottleneck distance between persistence diagrams. 
Informally (see Appendix, Section~\ref{appendix:hks} for a formal definition), it means that the points of $\Dg(G,t)$ must move smoothly with respect to $t$. 
This is experimentally illustrated in \cref{fig:hks_evol}, where we plot the four diagrams built from a graph of the \texttt{MUTAG} dataset.

\begin{figure*}
\centering
  \includegraphics[width=.9\textwidth]{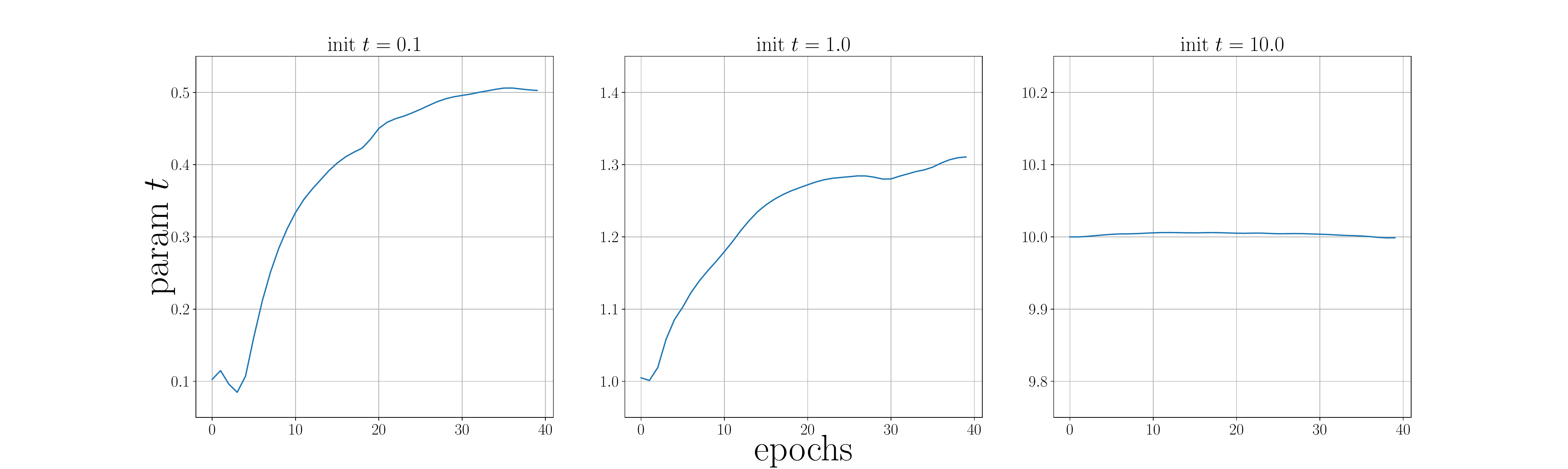}
  \caption{Evolution of HKS parameter $t$ when considered as a trainable variable (i.e.~differentiating $t \mapsto \Dg(G,t)$ for all $G$) across 40 epochs for three different initializations of $t$, namely 0.1, 1 and 10, on the \texttt{MUTAG} dataset.}
  \label{fig:auto_select_t}
\end{figure*}

As mentioned in \cref{subsec:hks}, the parameter $t$ can also be treated as a trainable parameter that is optimized during the learning. 
In our experiment, however, it does not prove to be worth it.
Indeed, our diagrams are not particularly sensitive to the choice of $t$, and thus fixing some $t$ sampled in log-scale is enough.
\cref{fig:auto_select_t} illustrates the evolution of parameter $t$ over 40 epochs when trained on the \texttt{MUTAG} dataset (one epoch correspond to a stochastic gradient descent performed on the whole dataset). 
As one can see, parameter $t$ converges quickly.
More importantly, it remains almost constant when initialized at $t_0 = 10.0$, suggesting that this choice is a (locally) optimal one.
Fortunately, this is the parameter we use in our experiment (see \cref{tab:expe_settings}).
On the other hand, each time $t$ is updated (that is, at each epoch), one must recompute the diagrams for all the graphs in the training set, significantly increasing the running time of the algorithm.

\subsection{Experimental settings}
\begin{table*}[ht]
\begin{center}
\resizebox{.7\textwidth}{!}{%
\begin{tabular}{|l|cc|cc|}
\cline{1-5}
	Dataset & Func.~used & PD preproc. & \textsc{PersLay} & Optim. \\
\cline{1-5}
	$\texttt{ORBIT5K}$      & $\mathrm{Alpha}_0$, $\mathrm{Alpha}_1$ & prom(500) & Pm(25,25,10,top-5)           & adam(0.01, 0., 300)     \\
	$\texttt{ORBIT100K}$    & $\mathrm{Alpha}_0$, $\mathrm{Alpha}_1$ & prom(500) & Pm(25,25,10,top-5)           & adam(0.01, 0., 300)     \\
\cline{1-5}
	$\texttt{REDDIT5K}$     & \hks{1.0}           & prom(500) & Pm(25,25,10,sum)           & adam(0.01, 0.99, 500)     \\
	$\texttt{REDDIT12K}$    & \hks{1.0}           & prom(500) & Pm(5,5,10,sum)             & adam(0.01, 0.99, 1000)    \\
	$\texttt{COLLAB}$       & \hks{0.1}, \hks{10} & prom(500) & Pm(5,5,10,sum)             & adam(0.01, 0.9, 1000)     \\
	$\texttt{IMDB-B}$       & \hks{0.1}, \hks{10} & prom(500) & Im(20,(10,2),20,sum)       & adam(0.01, 0.9, 500)     \\
	$\texttt{IMDB-M}$       & \hks{0.1}, \hks{10} & prom(500) & Im(10,(10,2),10,sum)       & adam(0.01, 0.9, 500)     \\
	$\texttt{COX2}$         & \hks{0.1}, \hks{10} & ---       & Im(20,(10,2),20,sum) & adam(0.01, 0.9, 500) \\
	$\texttt{DHFR}$         & \hks{0.1}, \hks{10} & ---       & Im(20,(10,2),20,sum) & adam(0.01, 0.9, 500) \\
	$\texttt{MUTAG}$        & \hks{10}            & ---       & Im(20,(10,2),10,sum) & adam(0.01, 0.9, 100) \\
	$\texttt{PROTEINS}$     & \hks{10}            & prom(500) & Im(15,(10,2),10,sum) & adam(0.01, 0.9, 70) \\
	$\texttt{NCI1}$         & \hks{0.1}, \hks{10} & ---       & Pm(25,25,10,sum)     & adam(0.01, 0.9, 300) \\
	$\texttt{NCI109}$       & \hks{0.1}, \hks{10} & ---       & Pm(25,25,10,sum)     & adam(0.01, 0.9, 300) \\

\cline{1-5}
\end{tabular}}
\end{center}
\caption{Settings used to generate our experimental results.}
\label{tab:expe_settings}
\end{table*}

\begin{table*}
	\begin{center}
		\resizebox{\textwidth}{!}{%
			\begin{tabular}{cc|c|c|c|c|c|c||c|c|c||c|c|}
				\cline{3-13}
				&                & \multicolumn{6}{c||}{\textbf{Grid size for trainable weights} $\mathbf{w(p)}$} & \multicolumn{3}{c||}{\textbf{Point transformation} $\mathbf{\phi}$} & \multicolumn{2}{c||}{\textbf{Perm op}} 
				\\ \cline{3-13} 
				&                & \texttt{None}  & $2\times 2$& $5 \times 5$&$10 \times 10 $&$20 \times 20$&$50 \times 50$& Gaussian & line & triangle & \texttt{Sum}          & \texttt{Max}     
				\\ \thickhline
				\multicolumn{1}{|c|}{\multirow{2}{*}{MUTAG}} & Train/Test acc (\%) & 92.3/88.9  & 91.1/88.8 & 91.7/89.6 & 92.3/\textbf{\textcolor{blue}{89.9}} & 93.7/88.3 & 94.1/87.7 & 92.5/\textcolor{blue}{\textbf{89.7}} & 89.2/84.2 & 91.5/85.0 & 92.3/\textcolor{blue}{\textbf{89.5}} & 91.9/87.4 
				\\ \cline{2-13} 
				\multicolumn{1}{|c|}{}                       & Run time, CPU (s)   &  2.30  & 2.77 & 2.79 & 2.77 & 2.77 & 2.78 & 2.80 & 5.91 & 4.42 & 2.75 & 2.82 
				\\ \thickhline
				\multicolumn{1}{|c|}{\multirow{2}{*}{COLLAB}}  & Train/Test acc (\%) & 76.5/75.3 &  78.6/75.8 & 79.0/76.2 & 80.0/\textcolor{blue}{\textbf{76.5}} & 83.5/73.9 & 94.0/71.3 & 79.7/75.3 & 79.9/\textcolor{blue}{\textbf{76.1}} & 79.4/74.7 & 80.0/\textcolor{blue}{\textbf{76.4}} & 78.8/75.0 
				\\ \cline{2-13} 
				\multicolumn{1}{|c|}{}                       & Run time, GPU (s) & 26.0 & 40.4 & 43.5 & 43.8 & 44.1 & 45.6 & 45.8 & 54.0 & 61.4 & 44.3 & 48.1 
				\\ \thickhline 
		\end{tabular}}
	\end{center}
	\vskip-0.35cm
	\caption{\emph{Influence of hyper-parameters and ablation study.} When varying a single hyper-parameter (e.g.~grid size), all the others (e.g.~perm op) are fixed to the values described in Supplementary Material, \cref{tab:expe_settings}. Accuracies and running times are averaged over 100 runs (i5-8350U 1.70GHz CPU for the small MUTAG dataset, P100 GPU for the large COLLAB one). Bold-blue font refers to the experimental setting used in \cref{sec:expe}.}
	\label{tab:influence_grid_size}
\end{table*}

Input data was fed to the network with mini-batches of size 128. For each dataset, various parameters are given (extended persistence diagrams, neural network architecture, optimizers, etc.) that were used to obtain the scores from \cref{tab:tableau_res}.
In \cref{tab:expe_settings}, we use the following shortcuts:
\begin{itemize}
	\item $\mathrm{Alpha}_d$: persistence diagrams obtained with $\texttt{Gudhi}$'s $d$-dimensional $\texttt{AlphaComplex}$ filtration.
	\item \hks{t}: extended persistence diagram obtained with HKS on the graph with parameter $t$.
	\item prom($k$): preprocessing step selecting the $k$ points that are the farthest away from the diagonal.
	\item \textsc{PersLay} channel Im($p$, ($a$, $b$), $q$, $\texttt{op}$) stands for a function $\phi$ obtained by using a Gaussian point transformation $\phi_\Gamma$ sampled on $(p \times p)$ grid on the unit square followed by a convolution with $a$ filters of size $b \times b$, for a weight function $w$ optimized on a $(q \times q)$ grid and for an operation $\texttt{op}$.
	\item \textsc{PersLay} channel Pm($d_1$, $d_2$, $q$, $\texttt{op}$) stands for a function $\phi$ obtained by using a line point transformation $\phi_L$ with $d_1$ lines followed by a permutation equivariant function~\cite{Zaheer2017} 
    in dimension $d_2$, for a weight function $w$ optimized on a $(q \times q)$ grid and for an operation $\texttt{op}$.
	\item adam($\lambda, d, e$) stands for the ADAM optimizer~\cite{Kingma2014} with learning rate $\lambda$, using an Exponential Moving Average\footnote{\url{https://www.tensorflow.org/api_docs/python/tf/train/ExponentialMovingAverage}} with decay rate $d$, and run during $e$ epochs. 
\end{itemize}

\subsection{Hyper-parameters influence}

As our approach mix our topological features and some standard graph features, we provide two ablations studies. In \cref{tab:complementary_expe}, the column ``Spectral'' reports the test accuracies obtained by using only these additional features, while the column ``PD alone'' records the accuracies obtained with the extended and ordinary persistence diagrams alone.
As ordinary persistence only encodes the connectivity properties of graphs, a gap in performance between extended and ordinary persistence can be interpreted as 1-dimensional features (i.e.~loops) being informative for classification purpose.
It also reports the standard deviations, that were omitted in \ref{tab:tableau_res} for the sake of clarity.

Similarly, we give in \cref{tab:influence_grid_size} the influence of the grid size that we choose as weight function $w$. 
In particular, we also perform an ablation study: grid size being \texttt{None} meaning that we enforce $w(p) = 1$ for all $p$.
As expected, increasing the grid size improves train accuracy but leads to overfitting for too large values. 
However, this increase has only a small impact on running times whereas not using any grid significantly lowers it.

\begin{figure}
	\begin{center}
		\includegraphics[width=0.9\columnwidth]{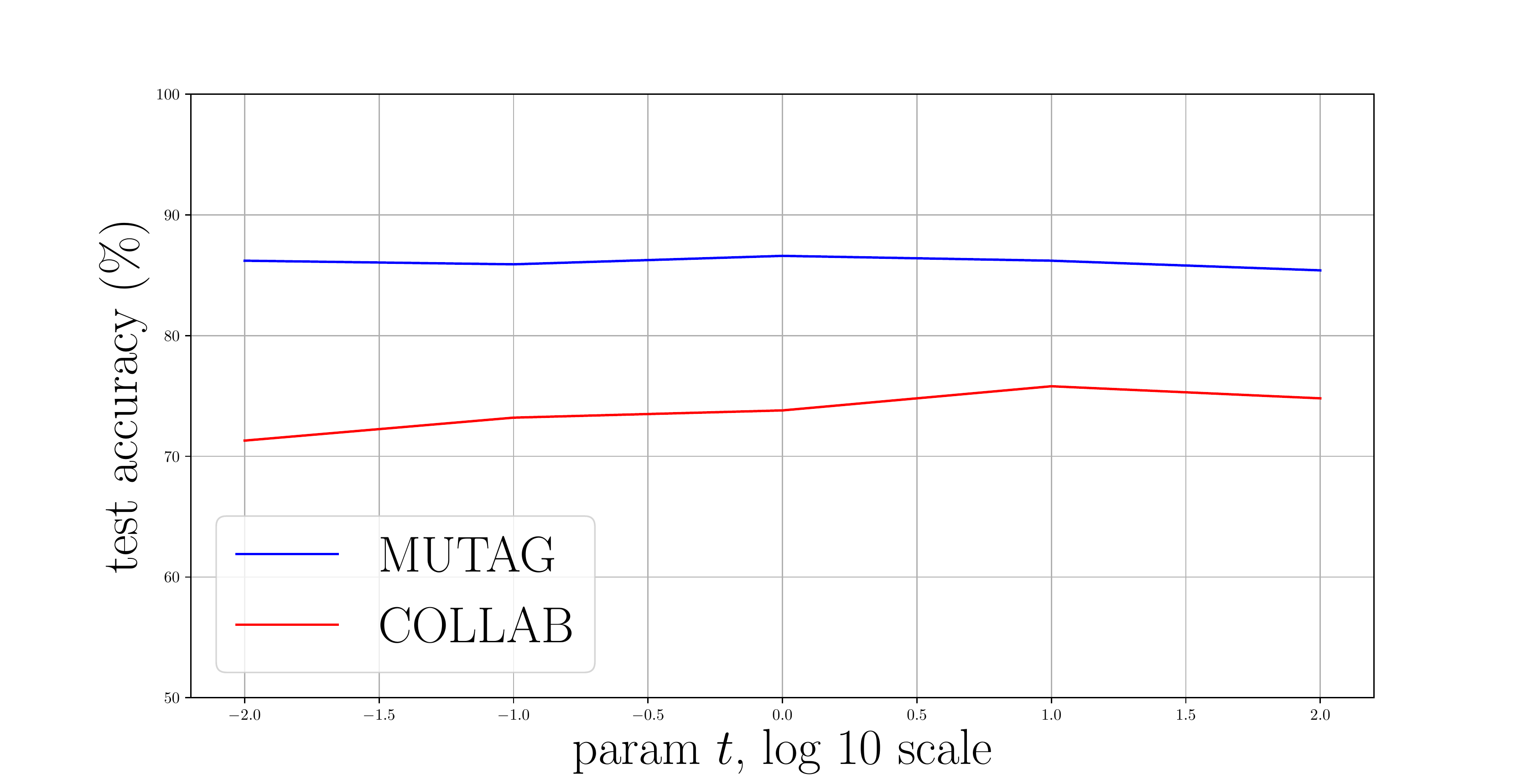}
	\end{center}
	\caption{Variation of test accuracy for \texttt{MUTAG} and \texttt{COLLAB} dataset when varying HKS parameter $t$ between $10^{-2}$ and $10^2$ (log-10 scale).}
	\label{fig:accuracy_wrt_t}
\end{figure}

Finally, \cref{fig:accuracy_wrt_t} illustrates the variation of accuracy for both \texttt{MUTAG} and \texttt{COLLAB} datasets when varying the HKS parameter $t$ used when generating the extended persistence diagrams. One can see that the accuracy reached on \texttt{MUTAG} does not depend on the choice of $t$, which could intuitively be explained by the small size of the graphs in this dataset, making the $t$ parameter not very relevant. Experiments are performed on a single 10-fold, with 100 epochs. Parameters of \textsc{PersLay} are set to \texttt{Im(20,(),20)} for this experiment.

\begin{table}[ht!]
	\begin{center}
		\resizebox{\columnwidth}{!}{%
			\begin{tabular}{|l||cccc|}
				\cline{1-5}
				& Spectral alone & \multicolumn{2}{c}{PD alone} & \textsc{PersLay} \\
				&                & Extended & Ordinary           &              \\
				\cline{1-5}
				$\texttt{REDDIT5K}$        & 49.7($\pm$0.3) & 55.0 & 52.5 & 55.6($\pm$0.3) \\
				$\texttt{REDDIT12K}$       & 39.7($\pm$0.1) & 44.2 & 40.1 & 47.7($\pm$0.2) \\
				$\texttt{COLLAB}$          & 67.8($\pm$0.2) & 71.6 & 69.2 & 76.4($\pm$0.4) \\
				$\texttt{IMDB-B}$          & 67.6($\pm$0.6) & 68.8 & 64.7 & 71.2($\pm$0.7) \\
				$\texttt{IMDB-M}$          & 44.5($\pm$0.4) & 48.2 & 42.0 & 48.8($\pm$0.6) \\
				$\texttt{COX2}$ \bf{*}     & 78.2($\pm$1.3) & 81.5 & 79.0 & 80.9($\pm$1.0) \\
				$\texttt{DHFR}$ \bf{*}     & 69.5($\pm$1.0) & 78.2 & 71.8 & 80.3($\pm$0.8) \\
				$\texttt{MUTAG}$ \bf{*}    & 85.8($\pm$1.3) & 85.1 & 70.2 & 89.8($\pm$0.9) \\
				$\texttt{PROTEINS}$ \bf{*} & 73.5($\pm$0.3) & 72.2 & 69.7 & 74.8($\pm$0.3) \\
				$\texttt{NCI1}$ \bf{*}     & 65.3($\pm$0.2) & 72.3 & 68.9 & 73.5($\pm$0.3) \\
				$\texttt{NCI109}$ \bf{*}   & 64.9($\pm$0.2) & 67.0 & 66.2 & 69.5($\pm$0.3) \\
				\cline{1-5}
			\end{tabular}
		}
	\end{center}
	\caption{Complementary report of experimental results.}
	\label{tab:complementary_expe}
\end{table}

\end{document}